\documentclass[sigconf,authorversion]{acmart}
\usepackage{multirow}
\usepackage{amsmath}
\usepackage{threeparttable}
\usepackage{adjustbox}
\usepackage{subfig}
\usepackage{float}

\usepackage{mathtools}
\AtBeginDocument{%
  }
\renewcommand\footnotetextcopyrightpermission[1]{}
\setcopyright{acmlicensed}
\copyrightyear{2024}
\acmYear{2024}
\acmDOI{XXXXXXX.XXXXXXX}

\acmConference[Conference acronym 'MM]{ACM Multimedia 2024 Workshop BCI}{June 03--05,
  2024}{Woodstock, NY}
\acmISBN{978-1-4503-XXXX-X/18/06}
\acmSubmissionID{6}
\begin{document}

\title{Towards Linguistic Neural Representation Learning and Sentence Retrieval from Electroencephalogram Recordings}
\author{Jinzhao Zhou}
\email{jinzhao.zhou@uts.edu.au}
\affiliation{
  \institution{}
  \city{}
  \country{}
}


\author{Yiqun Duan}
\email{yiqun.duan-1@uts.edu.au}
\affiliation{
  \institution{}
  \city{}
  \country{}
}

\author{Ziyi Zhao}
\email{ziyi.zhao-2@student.uts.edu.au}
\affiliation{
  \institution{}
  \city{}
  \country{}
}

\author{Yu-Cheng Chang}
\email{fred.chang@uts.edu.au}
\affiliation{
  \institution{}
  \city{}
  \country{}
}

\author{Yu-Kai Wang}
\email{yukai.wang@uts.edu.au}
\affiliation{
  \institution{}
  \city{}
  \country{}
}

\author{Thomas Do}
\email{thomas.do@uts.edu.au}
\affiliation{
  \institution{}
  \city{}
  \country{}
}

\author{Chin-Teng Lin}
\email{chin-teng.lin@uts.edu.au}
\affiliation{
  \institution{}
  \city{}
  \country{}
}
\renewcommand{\shortauthors}{Zhou et al.}
\begin{abstract} 
Decoding linguistic information from non-invasive brain signals using EEG has gained increasing research attention due to its vast applicational potential. Recently, a number of works have adopted a generative-based framework to decode electroencephalogram (EEG) signals into sentences by utilizing the power generative capacity of pretrained large language models (LLMs). However, this approach has several drawbacks that hinder the further development of linguistic applications for brain-computer interfaces (BCIs). Specifically, the ability of the EEG encoder to learn semantic information from EEG data remains questionable, and the LLM decoder's tendency to generate sentences based on its training memory can be hard to avoid. These issues necessitate a novel approach for converting EEG signals into sentences. In this paper, we propose a novel two-step pipeline that addresses these limitations and enhances the validity of linguistic EEG decoding research. We first confirm that word-level semantic information can be learned from EEG data recorded during natural reading by training a Conformer encoder via a masked contrastive objective for word-level classification. To achieve sentence decoding results, we employ a training-free retrieval method to retrieve sentences based on the predictions from the EEG encoder. 
Extensive experiments and ablation studies were conducted in this paper for a comprehensive evaluation of the proposed approach. Our evaluation results demonstrate that our EEG encoder achieves up to 55.15\% top-20 classification accuracy with unseen EEG signals. Visualization of the top prediction candidates reveals that our model effectively groups EEG segments into semantic categories with similar meanings, thereby validating its ability to learn patterns from unspoken EEG recordings. Additionally, using the predicted classification results, our retrieval method attains a recall@5 of up to 55.55\% and a BLEU-1 score of 30.44\% for sentence-level evaluation. Despite the exploratory nature of this work, these results suggest that our method holds promise for providing more reliable solutions for converting EEG signals into text.
\end{abstract}


\received{20 July 2024}
\received[revised]{12 March 2009}
\received[accepted]{5 June 2009}

\maketitle
\section{Introduction}
Decoding linguistic information from brain signals has traditionally relied on intracranial approaches, which offer promising prospects for restoring communication abilities in individuals with paralysis or spinal cord injuries~\cite{moses2021neuroprosthesis,willett2021high,metzger2022generalizable}. 
In contrast, the use of non-invasive brain signals such as EEG in linguistic decoding has only recently begun to attract research attention, due to their superior temporal resolution, portability, and safety~\cite{alarcao2017emotions,alday2019m,gaudet2020functional}. 

For its ability to measure surface neural activity with high temporal resolution and detect a diverser range rhythmic patterns, EEG signals can capture electrical activity in the sensorimotor cortex, which is known to produce $\mu$ rhythms rich in information during speech production~\cite{saltuklaroglu2018eeg}. This inherent connection between EEG signals and speech has led to various successful approaches in decoding EEG into linguistic units such as syllables~\cite{arjestan2016brain}, phonemes~\cite{cooney2019optimizing, cooney2018mel}, and words~\cite{gonzalez2017sonification, mohanchandra2016communication}, despite limitations due to the scale of available datasets and subject variability.

On the other hand, decoding sentences from EEG signals during unspoken reading tasks presents several unique challenges. Firstly, unspoken speech elicits less discriminative brain activity compared to spoken speech, making it harder to distinguish between different neural responses~\cite{proix2022imagined}. Second, there exists significant data sparsity, as the number of semantic categories is large while the dataset size remains relatively small~\cite{nieto2022thinking}. Third, the noisiness of thought during reading further complicates the decoding task. For instance, participants may not focus on every word equally, often paying less attention to grammatical words and more to words that contain crucial or interesting information from the sentence~\cite{duan2023dewave_brain2text}. Existing methodologies for decoding reading sentences from EEG signals have predominantly relied on a framework that pairs an EEG encoder with a pretrained large language model (LLM) decoder, training and decoding sentences by the machine translation approach~\cite{wang2022open_aaai_eeg2text,feng2023aligning}. However, recent analyses suggest that when training the EEG encoder with an LLM using a machine translation objective, the encoder's ability to genuinely learn to capture semantic EEG patterns remains unclear. Instead, the overpowered LLM decoder may generate sentences simply based on its training memory regardless of the EEG input~\cite{jo2024eeg}. These empirical findings underscore the necessity to validate the efficacy of learning EEG encoders from EEG reading tasks and highlight the limitations of using overly powerful pretrained LLMs for converting EEG signals into sentences. 

To overcome the aforementioned limitations, we aim to develop a novel approach for EEG-to-sentence conversion, which seeks to eliminate the bias introduced by the training memory of an overpowered LLM while enabling the assessment of the semantic information an EEG encoder learns from text-reading EEG data. To achieve this, we propose EEG-to-Text Retrieval (ETER), which consists of an EEG encoder and an unbiased sentence retrieval method. In particular, We first train our conformer-based EEG encoder using a masked contrastive learning loss to learn semantic EEG representations. Then we combine our EEG encoder with a classification head to predict a semantic keyword set (SK) for each input EEG signal. In the next stage, we employ a beam search retriever (BSR) to find the most relevant sentences based on the SK sets generated by our EEG encoder. Our two-step EEG-to-text retrieval method introduces two key features: it allows for a transparent evaluation of the EEG learning efficacy through word-level classification outputs and provides a training-free method to leverage these word-level results for sentence-level outputs. Extensive experiments and ablation analyses reveal that our EEG encoder effectively learns semantic EEG representations, achieving high accuracy in word-level classification. Additionally, results from the retriever demonstrate that the predictions from the first stage enable the retrieval of the correct ground-truth sentences. These findings underscore the feasibility of using a retrieval-based method for converting EEG signals into coherent sentences.

The main contributions of this paper are summarized as follows:
\begin{itemize}    
    \item We propose a novel retrieval-based approach for EEG-to-text conversion tasks. The ETER method leverages the output of a word-level EEG classifier to retrieve the most relevant sentence, thus eliminating the over-reliance on generative LLM decoders.
    \item We demonstrate the learning of effective semantic EEG representations using a Conformer-based EEG encoder trained with a masked contrastive objective. Visualization of the top prediction candidates further proves its capability to predict EEG signals as semantically related words.
    \item We designed a beam search retrieval method to efficiently retrieve relevant sentences from the prediction results of our Conformer-based EEG encoder. Despite the imperfect prediction from the EEG encoder, our retrieval method remains a viable solution for finding the correct sentence.
    \item We conduct extensive experiments to thoroughly validate the performance of the proposed ETER approach. Additionally, ablation studies confirm the vocabulary scalability of the method and validate our design choices, presenting a novel avenue for developing a linguistic BCI system.
\end{itemize}

\section{Related Works}
\noindent\textbf{Linguistic unit or word decoding from brain signals} 
Due to EEG's capacity to capture neural activities associated with speech production, pioneering words on linguistic decoding using EEG mainly focus on the decoding of linguistic units such as syllables or phenomes \cite{d2009toward,cooney2019optimizing, tamm2020classification}. For instance, \citep{brigham2010imagined} proposed to extract autoregressive coefficients as features for imagined syllable classification with a k-nearest neighbor (KNN) classifier. \citep{deng2010eeg} leveraged the Hilbert spectrum to extract features and classify the syllables using a Bayesian classifier. 

To decode higher-level semantics, numerous studies have dedicated efforts to word-level classification using EEG signal ~\cite{wang2013analysis,gonzalez2017sonification,zhao2015classifying,pawar2020multiclass,pawar2020multiclass,asghari2022eeg}. However, most of these studies have trained and evaluated their models on a very limited dataset, typically comprising only $4$ to $10$ words or a narrow set of directional words. As a result, recent research has sought to expand the output vocabulary scope to a more practical scale, either at the word-level~\cite{lopez2022state} or the pre-word level~\cite{willett2023high}. \citep{D_fossez_2023_meg_eeg_clip_pretrain_meta_brain2speech} used a large-scale word-level EEG dataset collected during listening, they enhance word-level classification accuracy through contrastive learning to align E/MEG signals with speech.\\

\noindent \textbf{End-to-End Decoding from EEG to sentence}
The recent trend in EEG-based sentence decoding on the other hand predominantly employs end-to-end machine translation approaches. For instance, EEG-to-Text~\cite{wang2022open_aaai_eeg2text} pioneered open-vocabulary decoding of EEG signals into sentences, establishing an initial performance benchmark. In their work a Transformer-based EEG encoder is used to transform EEG signals into EEG representations while a pre-trained LLM model takes these EEG representations as input and generate sentences. Building upon this, DeWave~\citep{duan2023dewave_brain2text} advanced decoding performance by introducing discrete codex and achieved text decoding directly from raw EEG waves. Subsequent innovations such as BELT~\cite{zhou2023belt,chau2023belt} and Curriculum Contrastive~\cite{feng2023aligning} introduced contrastive learning to enhance encoding quality. Additionally, NuSpeech~\cite{yang2024decode} leveraged the end-to-end speech decoding model Whisper~\cite{radford2023robust} to achieve commendable performance.

However, these end-to-end methods are prone to issues where a newly initialized EEG encoder combined with a powerful pretrained language decoder leads to the decoder merely memorizing and generating the training text without truly utilizing information from the EEG modality. Consequently, this may result in the EEG encoder failing to learn to capture EEG patterns. Diverging from these LLM-based approaches, our work first develops an effective EEG encoder for word-level classification and explores the feasibility of using a training-free, unbiased retrieval method to achieve sentence-level output. This approach eliminates potential limitations such as implicit teacher forcing evaluation or test sentence leakage.

\section{EEG-to-Text Retrieval}
\begin{figure}[h]
  \centering
  \includegraphics[width=0.9\linewidth]{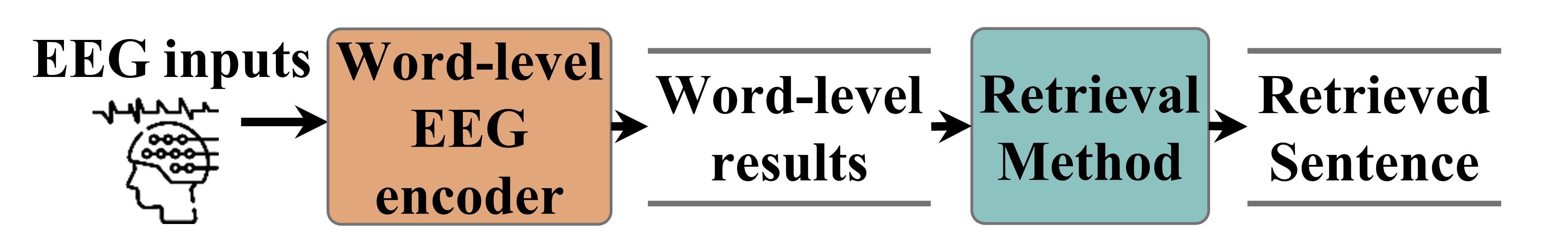}
  \caption{The overall structure of EEG-to-text retrieval approach. Our two-step approach consists of an word-level EEG encoder that encodes and prediction word-level results from EEG signals. Subsequently, a retrival method is applied to find the most relevant sentence utlizing the word-level results. \label{fig:cover}}
\end{figure}

\begin{figure*}[h]
  \centering
  \includegraphics[width=0.9\linewidth]{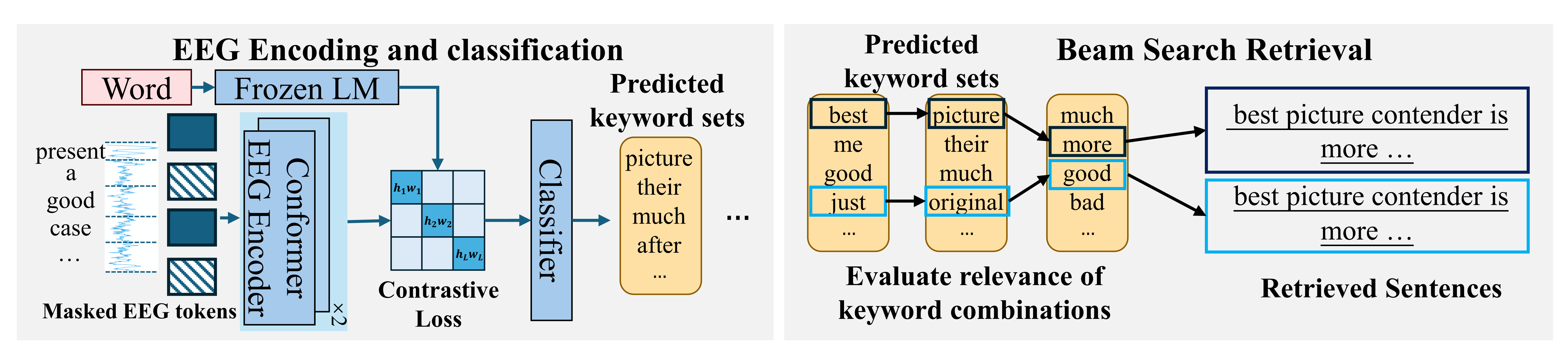}
  \caption{After segmenting and tokenizing EEG signals at the word level, an EEG encoder comprising two Conformer blocks learns semantic patterns from the EEG signals. We implement masked contrastive learning, leveraging a frozen language model to provide language supervision to the EEG representations. The EEG encoder predicts a keyword set independently for each input EEG segment. Finally, a sentence retriever utilizes these predicted keyword sets to identify the most pertinent sentence from the dataset corpus.\label{fig:pipeline}}
\end{figure*}

In this section, we present our ETER approach, a two-step EEG-to-text retrieval method that identifies the most relevant sentence a participant reads based on word-level EEG classification results. The general pipeline of our approach is depicted in Figure \ref{fig:cover}. To achieve word-level decoding, we developed a Conformer-based EEG encoder. To learn semantic EEG representation, we guide the EEG representation space using word representations extracted from a large language model and train the encoder with a masked contrastive objective. It is important to note that the language model is used solely to provide word representations for guiding the distribution of the learned EEG representation and is not utilized during testing. Subsequently, a classification head is added to the EEG encoder to fine-tune it for word-level prediction. Finally, we designed a beam search retrieval method to find relevant sentences based on the prediction results of the EEG encoder. Figure \ref{fig:pipeline} gives an overview of the proposed method. 

\subsection{Preprocessing}\label{sec:preprocess} 
To perform word-level EEG representation learning and classification, we first preprocess the dataset's vocabulary. Grammatical words such as "the," "a," "an," and "is" constitute a significant portion (40-60\%) of English text in general~\cite{kontra1985frequency, leech2014word}. From a sample balance perspective, these grammatical words dominate the training and testing samples, potentially leading the EEG encoder or classifier to overemphasize on these words, which do not contain critical information about the sentence. Furthermore, previous neurobiological studies in reading comprehension has identified that ``semantic strong'' words elicit higher and more distinguishable neural patterns compared to ``semantic moderate'' words~\cite{kutas1980reading}. Therefore, during preprocessing, we remove EEG-word pairs containing these grammatical words from the dataset.

Additionally, we perform word lemmatization on the remaining vocabulary. The lemmatization step serves two purposes. First, we hypothesize that during reading comprehension, different forms of the same word will elicit similar neural patterns, as they convey the same meaning. So the EEG signals for these similar words can be seen as the same category. Second, this lemmatization process also increases the sample size for each word in the vocabulary and reduces the sparsity of the word-level training dataset.

For preprocessing the EEG signals, they are first transformed into word-level EEG embeddings using frequency-domain transformation following the same preprocessing pipeline in previous works~\cite{hollenstein2018zuco,wang2022open}. First, the EEG recordings are segmented according to the eye-tracking fixation on each word. Then, the segmented EEG signals are denoised and band-pass filtered into eight frequency bands: theta1 (4-6Hz), theta2 (6.5-8Hz), alpha1 (8.5-10Hz), alpha2 (10.5-13Hz), beta1 (13.5-18Hz), beta2 (18.5-30Hz), gamma1 (30.5-40Hz), and gamma2 (40-49.5Hz). The Hilbert transform is then applied to each channel. Finally, word-level EEG embeddings are obtained by averaging the frequency band power within each frequency band. In the remainder of this paper, we denote the word-level EEG embedding as $\mathbf{e}$. For the corresponding word of the EEG embedding, we use the embedding layer of a distilled BERT model~\cite{devlin2018bert} to convert it into word representation, denoted by $\mathbf{w}$ for brevity. To enhance word-level EEG representation learning and classification performance, we apply standard normalization to the word-level EEG embeddings. Specifically, we compute the mean and standard deviation of $\mathbf{e}$ for each subject and use these values for applying standard normalization. Empirically, we found that this normalization stabilizes the training process and improves performance, likely by suppressing noise and reducing inter-subject variations to some extent.

\subsection{EEG Encoder}
We train an EEG encoder for encoding and classifying EEG signals. We first tokenize $\mathbf{e}$ into frequency tokens and then feed them to a Conformer encoder. The Conformer encoder outputs the same number of tokens as input, we use a global pooling layer to aggregate the information across all frequency bands into the final EEG representation $\mathbf{h}$.

\subsubsection{Frequency-wise EEG tokenization}
After preprocessing, the word-level EEG embedding has the shape of $\mathbf{e}\in\mathbb{R}^{N\times{D}}$. Here, $N$ denotes the number of channels, and $D$ is the number of frequency bands (in our case $D=8$). To tokenzie the EEG, we split $\mathbf{e}$ into non-overlapping frequency bands across all channels $\{\mathbf{e}^{(i)}\}_{i=1,\cdots, D}$. Since these EEG channels are distributed spatially on a participant's head so we employ spatial operations here to capture and aggregate frequency responses in a specific scalp area. As depicted in Figure \ref{fig:encoder}, we use a spatial encoder to transform $\mathbf{e}^{(i)}$ into EEG token. The spatial encoder consist of a lightweight convoutional network. The spatial encoder comprises a lightweight convolutional network and a positional embedding layer. The convolutional network processes the channel dimension to produce embeddings that consolidate spatial information from specific frequency bands. Concurrently, the positional embedding layer is used to encode the positional information of the frequency bands, indicating which frequency range is contained within the input $\mathbf{e}^{(i)}$. 

\subsubsection{Conformer for EEG encoding}
\begin{figure}[H]
  \centering
  \includegraphics[width=0.8\columnwidth]{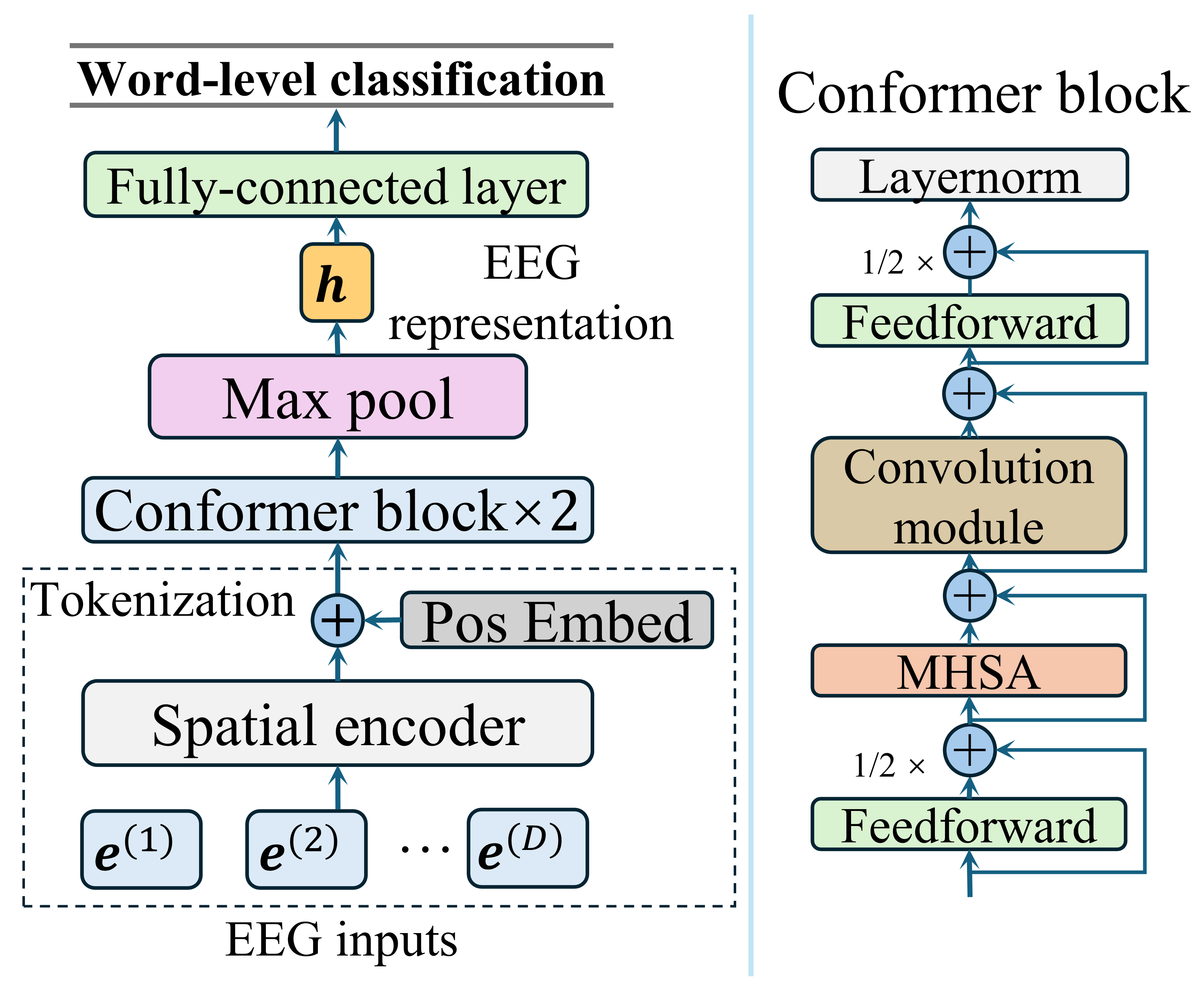}
  \caption{The architecture of the proposed EEG encoder. We first tokenize the low-level representations of EEG segments using a spatial encoder on the channel dimension and add positional embedding to indicate the frequency range of the token. Then we use $2$ conformer blocks to further process these tokens. The Conformer block encodes input tokens using a multi-head self-attention (MHSA) and a convolution module. For classification, we use an adaptive max pooling layer to aggregate the output of all EEG tokens into the final EEG representation $\mathbf{h}$ for each word. For classification, a fully-connected layer will be used to the classification distribution of the input EEG signal.\label{fig:encoder}}
\end{figure}

The detailed architecture of our EEG encoder is depicted in Figure \ref{fig:encoder}. We use the conformer blocks~\cite{gulati2020conformer} to build our EEG encoder for capturing both spectrum dependencies across EEG frequency bands and spatial relationships among channels~\cite{zhou2023belt,song2022eeg}. To aggregate the encoded EEG representations across all frequency bands, we used a global adaptive pooling layer to the output of the last confomer block and outputs $\mathbf{h}$ as the final EEG representation for each word. 

\begin{figure}[H]
  \centering
  \includegraphics[width=0.8\linewidth]{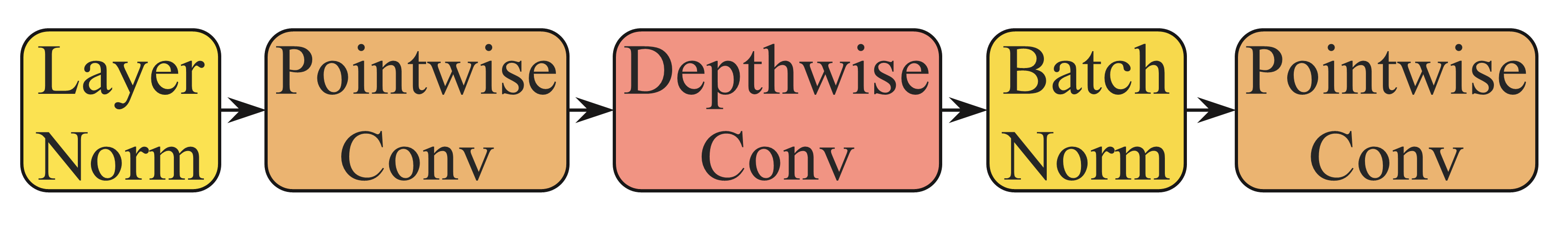}
  \caption{The detailed structure of the convolution module used in the Conformer blocks.\label{fig:convolution-module}}  
\end{figure}

In a Conformer block, two feed-forward networks ($F\!F\!N^1$ and $F\!F\!N^2$), a multi-head self-attention (MHA) module, a convolution (CN) module are stacked together using residual connections. We applied a $1/2$ weigh for the two $F\!F\!N$ layers. The convolution module is depicted in Fig. \ref{fig:convolution-module}, which is in turn comprised of two pointwise convolution layers and a depthwise convolution layer. The first pointwise convolution layer of the convolution module uses the gated linear unit (GLU) as the activation function. A batch normalization layer and a swish activation function were also used after the depthwise convolution layer. Overall, the Comformer blocks take the EEG embeddings $\mathbf{e}$ as input and output the continuous EEG representation $\mathbf{h}$.

\subsubsection{Masked contrastive training} 
\begin{figure}[H]
  \centering
  \includegraphics[width=0.8\columnwidth]{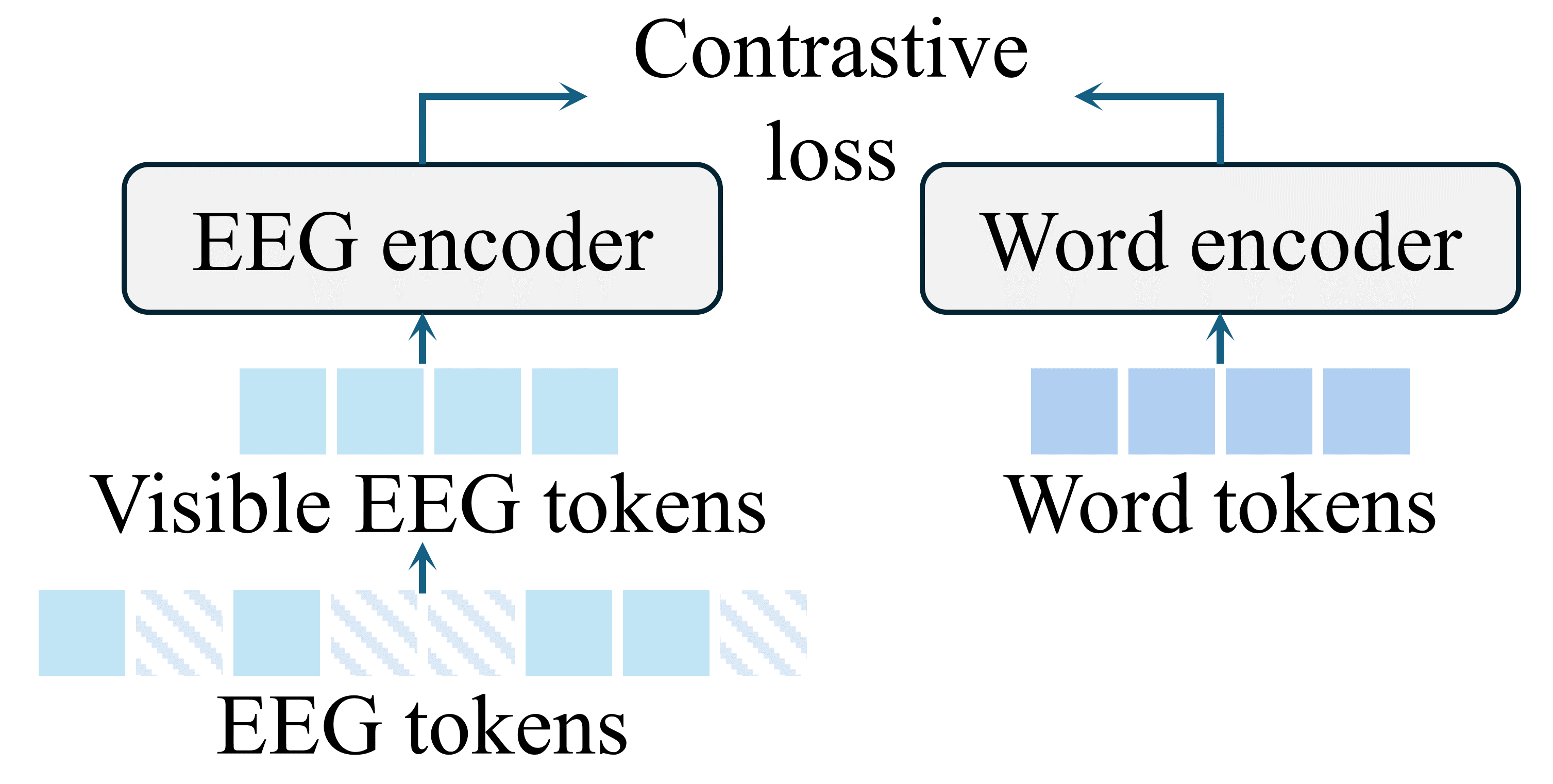}
  \caption{Our masked contrastive training scheme. Following CLIP~\cite{radford2021learning} and FLIP~\cite{li2023scaling}, we perform contrastive learning on pairs of EEG segments and words. We randomly mask out EEG tokens with a small masking ratio and encode only the visible EEG tokens. Here, the slash-shadowed blocks mean the EEG token is masked and is not visible to the EEG encoder. We do not mask word tokens in our paper. \label{fig:maskct}}
\end{figure}

To train the EEG encoder, we employ the masked contrastive learning objective, as depicted in Figure \ref{fig:maskct}. This self-supervised approach aligns EEG representations $\mathbf{h}$ with word representations $\mathbf{w}$, enabling the EEG encoder to extract semantic information from EEG signals. This alignment ensures that EEG representations are not only closely related to its groundtruth word category but also to words with similar meanings. To further enhance the robustness of the EEG representations, we apply random masking to the input EEG tokens with a masking ratio $\eta$. Notably, we do not apply masking to the word embeddings to avoid introducing unnecessary noise into the learning process. The masked contrastive training loss function is defined by $\mathcal{L}^{ct}$ as follows:

\begin{equation}\label{eq:contrastive}
\mathcal{L}^{ct} =-\frac{1}{M}\sum^M_{i=1} \log\frac{\exp{\rm{sim}(\mathbf{h}_i,\mathbf{w})_i/\tau}}{\sum^M_{j=1}\exp\rm{sim}(\mathbf{h}_i,\mathbf{w}_j)/\tau}
\end{equation}
, where $M$ is the training sample size of the dataset, $\tau$ is the temperature parameter that scales the logits, and $\text{sim}(\cdot, \cdot)$ denotes the dot product similarity measure. We employ a frozen, pretrained BERT model~\cite{devlin2018bert} as the text encoder to generate word representations and guide the learning of EEG representations. In our experiments, we empirically determined that a mask ratio of $\eta = 0.1$ and a temperature parameter of $\tau = 0.3$ yield optimal classification performance.

\subsubsection{Word-level classification} 
While training an EEG encoder with a self-supervised objective provides a robust foundation for learning semantic representations, it alone is insufficient for effective EEG classification. To address this limitation, we introduce a supervised learning phase that augments the self-supervised training with an additional classification head. We use a fully-connected layer with softmax activation function as the classification head using the EEG representation $\mathbf{h}$. This layer maps the language-aligned EEG representations to specific word categories, leveraging supervised loss $\mathcal{L}^{sup}$ (Equation \ref{eq:sup}) to refine the encoder's predictions.
\begin{equation}\label{eq:sup}
\mathcal{L}^{sup} = -\frac{1}{M}\sum^M_{i=1}y_i\log(p(\hat{y}_i|\mathbf{h}_i))
\end{equation}
, $M$ denotes the number of training samples, $y_i$ represents the one-hot encoded target word for the $i$-th sample, $hat{y}_i$ is the predicted word. In addition to the fully-connected layer classifier, we apply regularization techniques such as dropout and weight decay to prevent overfitting and ensure that the model generalizes well to unseen EEG samples.

\begin{figure*}[ht!]
  \centering
  \includegraphics[width=1\linewidth]{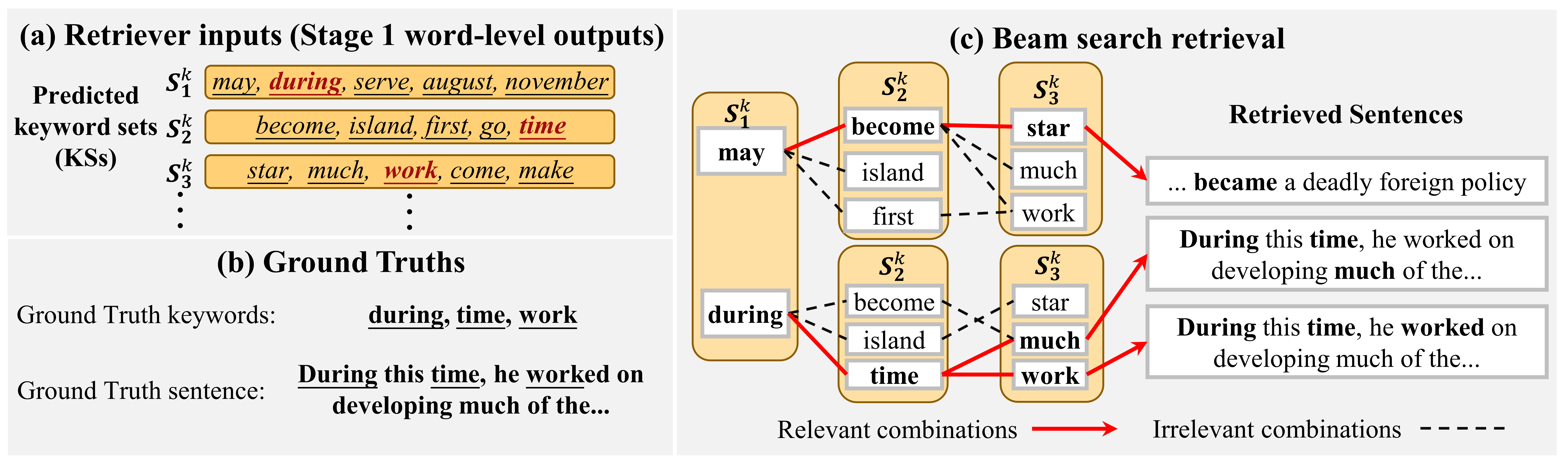}
  \caption{Illustration of the BSR method for reteiving reading sentences using word-level classification results from previous stage. (a) An example of the retriever's input. Although stage-1 model fails to predict the correct word in its top-1 prediction, it predicts correct words within the top-k KS. (b) Ground truth words  reading sentence for reference. (c) The BSR method retrieves the ground truth sentence using these KSs iteratively. \label{fig:beamsearch}}
\end{figure*}

\subsection{Sentence Retrieval Method} 
In this section, we introduce the retrieval method designed to achieve EEG-to-text conversion based on the results from our word-level classification model. Our word-level EEG encoding and classification approach, as introduced previously, provides a solid and transparent measure of how well the encoder captures linguistic patterns from EEG data by allowing the direct evaluation using accuracy metrics. However, achieving high top-1 accuracy in linguistic EEG classifications remains a significant challenge under a large vocabulary as reported in previous works~\cite{bakhshali2020eeg,rusnac2022imaginary,defossez2023decoding}. 

To address this limitation, we leverage a characteristic that emerged from our masked contrastive learning approach. After training, our model can generate top-$k$ word predictions with similar meanings from the input EEG signals. This capability is crucial as it mitigates the challenges of achieving precise top-1 classification by aggregating semantically related words. This aggregation enhances the robustness and accuracy of our retrieval method, allowing for more reliable decoding of EEG signals into meaningful sentences.
We denote the group of top-$k$ prediction words as a keyword set (KS), denoted by $\mathbf{S}^k$. Here, $k$ denotes the number of top prediction words. 
Building upon this, we design our retrieval method to leverage the $\mathbf{S}^k$ from each ``EEG word'' to identify the most relevant sentence from the reading corpus. We denot the sequance of KSs in a sentence as $\mathcal{S}=\{\mathbf{S}_i^k\}_{i=1,\cdots,L}$, where $L$ denotes the number of KS predicted for the sentence. 
\subsubsection{Beam search retrieval method}
We depict the proposed beam search retrieval method (BSR) in Figure \ref{fig:beamsearch}. The BSR method is designed to leverage a large search space that considers all $k$ candidates in $\mathbf{S}^k$, while reducing exponential memory consumption. BSR begins by constructing keyword combination queries from the first $n$ KSs. Each query contains one candidate from a $\mathbf{S}^k$, and will be scored according to its relevance to sentences in the dataset corpus. The scoring method will be explained in Section \ref{sec:scoring}. The score for each query measures the relevance of this query to the dataset corpus. After scoring, we apply re-ranking to the queries and only keep the best $m$ combination queries for the next evaluation round. In the next round, the $(n+1)^{th}$ KS will be added to the queries to produce further combination queries. This iterative method ensures that at each step, we maintain the most promising combinations, incrementally building up to the final sentence retrieval. Mathematically, this iterative beam search process can be described as follows:
\begin{equation}
\label{eq:beam}
\begin{split}
\mathbf{q}^0 &\coloneqq \{\emptyset\}\\
\mathbf{q}^l &= \mathop{\rm{argmax}}_{
        \mathbf{q}^\prime\subseteq\mathcal{B}_l
        \atop
        |\mathbf{q}^\prime|=m} H(\mathbf{q}^\prime,\mathcal{C})
\end{split}
\end{equation}
, here $\mathbf{q}^0$ denotes the initial combination query set before the interactive search. It is an empty set as there is no relevant query is kept at the start. $\mathbf{q}^l$ denotes the retained combination queries after the $l^{th}$ iteration. We use $\mathcal{B}_l$ to denote the new combinations obtained when adding the $l^{th}$ KS ($\mathbf{S}_l^k$) in this iteration. $H(\mathbf{q}^\prime,\mathcal{C})$ denotes a scoring method between the combinations $\mathbf{q}^\prime$ and sentences from the dataset corpus $\mathcal{C}$. We set $|\mathbf{q}^\prime|=m$ to limits the beam width of the searching. We calculate the candidate query set at $l>0$ by:
\begin{equation}
\mathcal{B}_l=\{q\circ{y}|q\subseteq\mathbf{q}^{l-1}, y\in KS_l^k \} 
\end{equation}
, where $\circ$ denotes the concatenation operation. 
We borrow the process depicted in Figure \ref{fig:beamsearch} as an illustrative example. Assume we have a total of $L=3$ KS in the sentence. Figure \ref{fig:beamsearch}(a) shows all KSs from stage 1. In this example, none of the KS predicted the ground truth word as its top-1 prediction. However, the correct word can be found within the top-$k$ prediction set. Figure \ref{fig:beamsearch}(b) illustrates the ground truth words and ground truth reading sentence for reference. The BSR method, as shown in Figure Figure \ref{fig:beamsearch}(c), compares a number of combinations to the dataset corpus, distinguishing relevant combinations from irrelevant ones. In our example, the relevant combinations are [may, become, star], [during, time, work], and [during, time, much]. Using these relevant combinations, our model is able to identify the closest sentences from the dataset, including the ground truth sentence "During this time, he worked..." and returns this as the retrieval result.

\subsubsection{Scoring Method}\label{sec:scoring}
We use the Aho-Corasick algorithm~\cite{aho1975efficient} as the training-free scoring method. In particular, the Aho-Corasick algorithm efficciently finds all occurences of the combination query within a sentence from the corpus by constructing a finite state machine. Thus, we denote the calculation of $H(\mathbf{q},\mathcal{C})$ by:
\begin{equation}
H(\mathbf{q},\mathcal{C}) = \sum_{top-m}\max{|q\cap{c}|}, c\in{\mathcal{C}},q\in\mathbf{q}, 
\end{equation}
, where $|q\cap{c}|$ denotes the number of occurrences of a query within a sentence $c$. We score a query using its average occurrence match with the sentence to allow the tolerance of ``wrong keywords'' in the query. 

\section{Experiment}

\subsection{Dataset}

\begin{figure}[h]
  \centering  \includegraphics[width=1.0\linewidth]{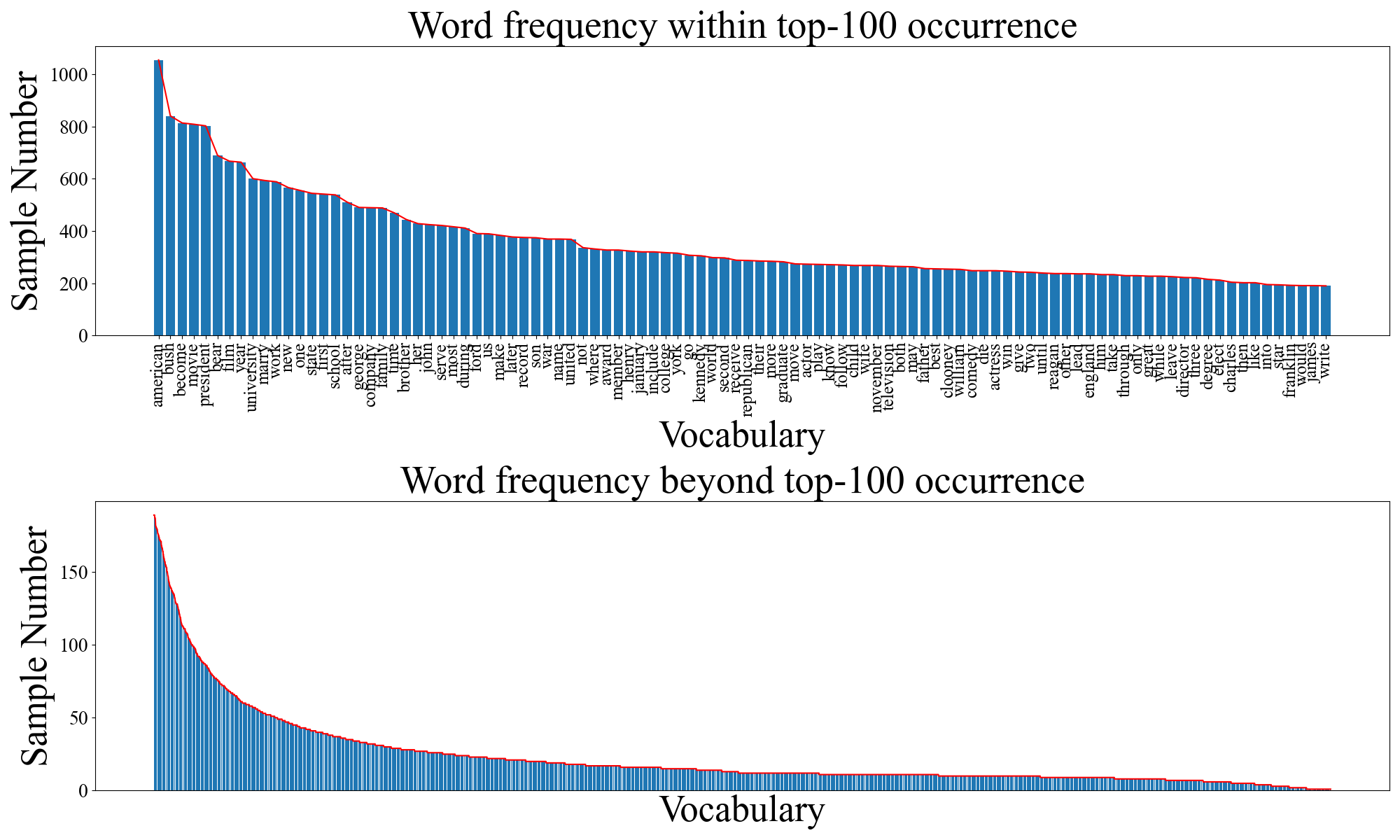}
  \caption{Samples number of the lemmatized vocabulary of the Zuco dataset. (top) sample number of words within the top 100 most occurring words. (bottom) sample number of words outside the top 100 most occurring words. A sharp decrease in sample number can be observed (red curve). 
  EEG-word preprocessing. We preprocess and lematize the original sentence from the dataset. EEG-word pairs are extracted for training the word-level EEG encoder. \label{fig:vocab-distribution}}
\end{figure}

In this study, we utilize the Zurich Cognitive Language Processing Corpus (ZuCo) dataset~\cite{hollenstein2018zuco} for training and evaluating the proposed method. The ZuCo dataset contains EEG data recorded during unspoken reading tasks involving 12 participants. It includes data from 105 EEG channels, with EEG waves denoised and filtered into eight frequency bands after segmentation. For our experiments, we use data from reading comprehension tasks, specifically Task 1 and Task 3 to evaluate the performance of our ETER method. Task 1 focuses on sentiment comprehension from movie reviews~\cite{socher2013recursive}, while Task 3 involves understanding and extracting entities' relation from Wikipedia biography articles. As discussed in Section \ref{sec:preprocess}, we removed all EEG-word pairs containing grammatical words from the dataset and performed lemmatization on the remaining words, merging words with the same lemmatized root form. Additionally, we observed a sharp decrease in sample numbers for words outside the top-100 most frequently occurring words in the remaining EEG-word pairs. As depicted in Figure \ref{fig:vocab-distribution}, most long-tailed cases have fewer than 30 samples in the entire dataset, with some extreme cases having only one sample. This imbalance problem results in significant sparsity in the training dataset. Making it impossible to develop any effective word-level models on the full vocabulary of the dataset. To address this issue, we selected only the 100 most frequently appearing words from the dataset for training our EEG encoder. Although this selection may limit the system's ability to scale, it provides relatively stable performance and serves as a reliable solution for our current needs. We have also conducted experiments involving a larger vocabulary in ablation studies, which will be discussed in Section \ref{sec:ablation-size}.

\subsection{Mectrics}
To ensure a thorough evaluation of our approach, we utilize a range of evaluation metrics for both the EEG classifier and the retrieval method. Firstly, we evaluate the effectiveness of the EEG encoder through classification accuracy assessment. In the context of sentence retrieval, we employ the retrieval mectics including recall@5 and precision@5 metrics to evaluate the ability of our system to retrieve relevant sentence based on the results from EEG classification. Additionally, we calculate the BLEU metric~\cite{brown2020language} to quantify the relevance between the retrieved sentences and the target sentence. 

\subsection{Implementation Details}
We train a Conformer encoder with $2$ Conformer blocks. We set the embedding dimension to $512$ with $8$ attention heads with the feed-forward dimension size of $1024$. During training, we set the coefficient for training loss as $\alpha=0.5$ and $\beta=0.5$ respectively. We optimize the parameters of the Conformer models using AdamW optimizer with an initial learning rate of $1e^{-4}$ and a weight decay of $0.05$. The learning rate warms up over the first $500$ steps to $1e^{-2}$ and linearly decays to $1e^{-6}$. In all experiments, we set the batch size to $256$ and train the model for $100$ epochs. Training is performed on a single A40 GPU with $48$ GBs of memory. 

\subsection{Word-level classification performance}
We train and evaluate our EEG encoder and its ablative versions using the ZuCo dataset to demonstrate its ability to learn semantic representations from unspoken EEG signals. For the baseline, we use a random model that predicts a uniform distribution over the EEG segments. Our initial model is a conformer EEG encoder trained solely with the supervised learning loss $\mathcal{L}^{sup}$, without subject-baseline removal. We then assess the performance gains by incorporating subject-baseline removal (+bm.) and masked contrastive loss (+MCT). As shown in Table \ref{tab:encoder}, our model predicts the correct word from EEG with a top-20 accuracy of 55.15\% and a top-10 accuracy of 36.4\%. This indicates that for more than half of the unseen EEG samples across different subjects, the ground truth words rank significantly higher than others within a 100-word vocabulary. Compared to the random baseline, our model achieves nearly three times higher accuracy. Furthermore, we observe that the addition of baseline removal and masked contrastive training improves the top-20 accuracy by $7.86\%$ and $7.62\%$, respectively. These results highlight the incremental improvement provided by these methods in learning linguistic EEG patterns during reading.
\begin{table}[h!]
\centering
\caption{Word-level classification accuracy (\%) on unseen EEG segments\label{tab:encoder}}
\begin{threeparttable}
\renewcommand\arraystretch{0.9}
\begin{tabular}{llllll}
\toprule
Method       & Top-1           & Top-5            & Top-10           & Top-15           & Top-20           \\ \midrule
Random model & 1.08\%          & 5.07\%           & 9.57\%           & 14.27\%          & 19.19\%          \\
base model (Ours)  & 5.31\%          & 16.36\%          & 26.11\%          & 32.72\%          & 39.67\%          \\
+ bm.        & 6.48\%          & 21.48\%          & 31.24\%          & 40.38\%          & 47.53\%          \\
+ MCT        & \textbf{8.66\%} & \textbf{24.90\%} & \textbf{36.40\%} & \textbf{46.28\%} & \textbf{55.15\%} \\
\bottomrule
\end{tabular}
\begin{tablenotes}
\item[1] bm. denotes baseline removal using standard normalization for each participant.
\item[2] MCT denotes masked constrastive training.
\end{tablenotes}    
\end{threeparttable}
\end{table}

\subsection{Sentence-level retrieval performance}
\begin{table*}[ht]
\centering
\caption{Sentence Retrieval Performance \label{tab:decoder}}
\begin{threeparttable}
\renewcommand\arraystretch{0.9}
\begin{tabular}{ll|llll|llll}
\toprule
Reading                    & Scoring & Recall@5  & Precision@5 & Recall@5  & Precision@5 & BLEU-1    & BLEU-4    & BLEU-1    & BLEU-4    \\
Corpus                     & Method  & ($T\ge5$) & ($T\ge5$)   & ($T\ge7$) & ($T\ge7$)   & ($T\ge5$) & ($T\ge5$) & ($T\ge7$) & ($T\ge7$) \\ \midrule
\multirow{3}{*}{(Task 1) Sentiment} & TF-IDF  & 2.94\%    & 0.73\%      & 0.20\%    & 0.06\%      & 6.20\%    & 0.00\%    & 9.58\%    & 0.00\%    \\
                           & L.D.    & 18.75\%   & 4.06\%      & \textbf{50.00\%}   & 10.00\%     & 14.60\%   & 1.85\%    & 18.03\%   & 3.47\%    \\
                           & A.C.    & \textbf{37.50\%}   & \textbf{22.90\%}     & \textbf{50.00\%}   & \textbf{19.64\%}     & \textbf{30.37\%}   & \textbf{20.68\%}   & \textbf{47.92\%}   & \textbf{41.81\%}   \\ \midrule
\multirow{3}{*}{(Task 3) Wikipedia} & TF-IDF  & 3.03\%    & 1.51\%      & 6.75\%    & 2.98\%      & 11.33\%   & 3.80\%    & 9.11\%    & 5.96\%    \\
                           & L.D.    & 7.69\%    & 2.10\%      & 15.00\%   & 4.58\%      & 16.21\%   & 4.72\%    & 21.35\%   & 8.20\%    \\
                           & A.C.    & \textbf{15.38\%}   & \textbf{5.55\%}      & \textbf{55.55\%}   & \textbf{20.37\%}     & \textbf{29.12\%}   & \textbf{14.08\%}   & \textbf{49.86\%}  & \textbf{36.27\%}   \\ \bottomrule
\end{tabular}
\begin{tablenotes}
\item[1] L.D. denotes Levenshtein distance.
\item[2] A.C. denotes Aho–Corasick score. 
\end{tablenotes} 
\end{threeparttable}
\end{table*}

We evaluate the performance of the second-stage retrieval method using the sentiment movie review corpus from Task 1 and the Wikipedia biography corpus from Task 3. For these evaluations, we impose constraints on the number of available words within the sentences, requiring at least 5 or 7 KSs, as shown in Table \ref{tab:decoder}. We compare the proposed BSR method using various scoring methods including the Aho-Corasick method, Levenshtein distance, and Term Frequency-Inverse Document Frequency (TF-IDF). When using Levenshtein distance, we compute the edit distance between the query and the compared sentences while for the TF-IDF method, we calculate the cosine similarity between the bag-of-word representations of the query and the comparison sentence retrieved from the corpus. As presented in Table \ref{tab:decoder}, our experiment demonstrates the superior performance of using the Aho-Corasick-based scoring with our BSR method to accurately retrieve relevant sentences from the corpus based on input keyword sets. In the sentiment movie review corpus, our method achieves a recall@5 metric of 37.5\% for sentences containing over 5 keyword sets. For sentences containing over 7 keyword sets in both corpora, we achieve a recall@5 of over 50\%. Moreover, our method demonstrates the highest performance in retrieving relevant sentences, as evidenced by the BLEU metrics, surpassing a BLEU-1 score of 40\% on both reading corpora for sentences with over 7 keyword sets. Since These results are achieved without requiring any training in the retrieval method, it showcase the plausibility of the proposed ETER method for EEG-to-text conversion. 

Aside from the quantitative results, Table \ref{tab:decoder-vis} presents a qualitative assessment of the proposed ETER method. For qualitative comparison with a generative LLM decoder, we additionally fine-tuned a T5 model~\cite{raffel2020exploring} to generate ground truth sentences using lists of keywords as input. We show that our approach effectively retrieved top-ranking sentences in the first example case. In contrast, the T5 model produced sentences outside the training dataset, which is largely based on its pre-training memory. In the last example, although our model failed to find the correct sentence, it still managed to successfully identify keywords like 'best', enabling retrieval of similarly sentiment sentences from the corpus. This underscores the efficacy of our retrieval-based method in transcribing EEG signals into text given an imperfect word-level EEG classifier.

\begin{table*}[ht!]
\centering
\small
\caption{Visualization of the retrieved sentences\label{tab:decoder-vis}}
\resizebox{0.9\linewidth}{!}{
\begin{tabular}{ll}
\toprule
GT Sentence    & \begin{tabular}[c]{@{}l@{}}hepburn win an emmy award in 19xx for her lead role in love among the ruin and be nominate for four\\  other emmy and two tony award during the course of her more than 70 year act career\end{tabular}          \\
T5 Generation  & \begin{tabular}[c]{@{}l@{}}Among those attending the Kennebunkport, Maine wedding ceremony were Isabel Stillman Rockefeller \\ (daughter of Percy Rockefeller), Hope Lincoln, Mary Keck...\end{tabular}                                     \\
ETER Query     & ['award', 'war', 'her']                                                                                                                                                                                                                     \\
ETER Retrieval & \textbf{\begin{tabular}[c]{@{}l@{}}hepburn win an emmy award in 19xx for her lead role in love among the ruin and be nominate for \\ four other emmy and two tony award during the course of her more than 70 year act career\end{tabular}} \\ \midrule
GT Sentence    & \begin{tabular}[c]{@{}l@{}}henry ford july 30 19xx    april 7 19xx be the founder of the henry ford motor company which   \\ later become cadillac and ford motor company\end{tabular}                                                      \\
T5 Generation  & During this period, McNamara helped to plan the 1945 bombing of Tokyo.                                                                                                                                                                      \\
ETER Query     & ['found', 'company', 'later', 'henry', 'ford', 'john',   'become', 'then']                                                                                                                                                                  \\
ETER Retrieval & \textbf{\begin{tabular}[c]{@{}l@{}}henry ford july 30 19xx april 7 19xx be the founder of the henry ford motor company which   \\ later become cadillac and ford motor company\end{tabular}}                                                \\ \midrule
GT Sentence    & like the best of godard's movie it be visually ravish   penetrate impenetrable                                                                                                                                                              \\
T5 Generation  & \begin{tabular}[c]{@{}l@{}}He was a member of the Executive Committee of the United States Golf Association (USGA) from \\ 1928-1935, serving successively as Secretary, Vice President and President.\end{tabular}                         \\
ETER Query     & ['other', 'best', 'not']                                                                                                                                                                                                                    \\
ETER Retrieval & \begin{tabular}[c]{@{}l@{}}it's the best film of the year so far the benchmark against which all other \\ best picture contender should be measure\end{tabular}                                                                             \\ \midrule
GT Sentence    & \textbf{adam receive one electoral vote in the presidential election of 19xx}                                                                                                                                                               \\
T5 Generation  & \begin{tabular}[c]{@{}l@{}}During this time, he was prescribed Ritalin for hyperactivity; years later, his wife Annette had \\ been prescribed Ritalin for hyperactivity;\end{tabular}                                                      \\
ETER Query     & ['adam', 'january', 'elect']                                                                                                                                                                                                                \\
ETER Retrieval & \begin{tabular}[c]{@{}l@{}}although adam lose in both the popular and electoral vote in the presidential \\ election of  19xx …\end{tabular}                                                                                     \\ \bottomrule
\end{tabular}
}
\end{table*}

\subsection{Ablations and Discussions}
\subsubsection{EEG encoder's architecture} 

We delve deeper into assessing the effectiveness of different encoder architectures using the same training paradigm as the proposed method. Specifically, we interchange Conformer blocks with Transformer blocks~\cite{vaswani2017attention} or Emformer blocks~\cite{shi2021emformer}, in our EEG encoder architecture and evaluate their word-level classification performance. Results are presented in Table \ref{tab:ablation-arch}. To begin with, both the Conformer and Emformer, with their ability to capture local patterns across channels, exhibit notably superior performance compared to the general Transformer encoder. This observation underscores the significance of leveraging structures that exploit local patterns, justifying our design choice of EEG encoder. Notably, the Conformer encoder achieves better performance than the Emformer-based encoder in our assessments, indicating that the convolutional layers in the conformer architecture allow the model to learn hierarchical features, which can be crucial for understanding complex signals such as EEG. Additionally, our results show that the introduction of a reconstructive term does not yield a consistence enhancement to the performance, further validating our choice of a masked contrastive learning scheme without a reconstructed decoder.

\begin{table*}[ht]
\caption{Word-level classification results from various architectures and encoding block selection. \label{tab:ablation-arch}}
\begin{tabular}{llllll}
\toprule
Encoding block   & Top-1           & Top-5            & Top-10           & Top-15       & Top-20           \\ \midrule
Transformer      & 5.66\%          & 19.11\%          & 29.23\%          & 35.89\%      & 40.01\%          \\
+ reconstruction & 5.16\% (-0.60)  & 18.03\% (-1.08)  & 26.60\%  (-2.63) & 32.51\%  (-3.38)    & 36.51\% (-3.50)  \\
Emformer         & 7.74\%          & 21.57\%          & 32.64\%          & 41.38\%      & 48.83\%          \\
+ reconstruction & 7.95\% (+0.21)  & 24.27\% (+2.70)  & 36.39\%  (+3.75) & 45.75\%  (+4.37)    & 53.75\% (+4.92) \\
Conformer        & \textbf{8.66\%} & \textbf{24.90\%} & 36.40\%          & \textbf{46.28\%}     & \textbf{55.15\%} \\
+ reconstruction & 8.48\% (-0.18)  & 24.72\%  (-0.18) & \textbf{36.95\%} (+0.55)    & 45.97\% (-0.31)  & 54.32\% (-0.83)          \\ \bottomrule
\end{tabular}
\end{table*}

\subsubsection{Vocabulary size}\label{sec:ablation-size}

\begin{figure}[ht]
  \centering
  \includegraphics[width=1\linewidth]{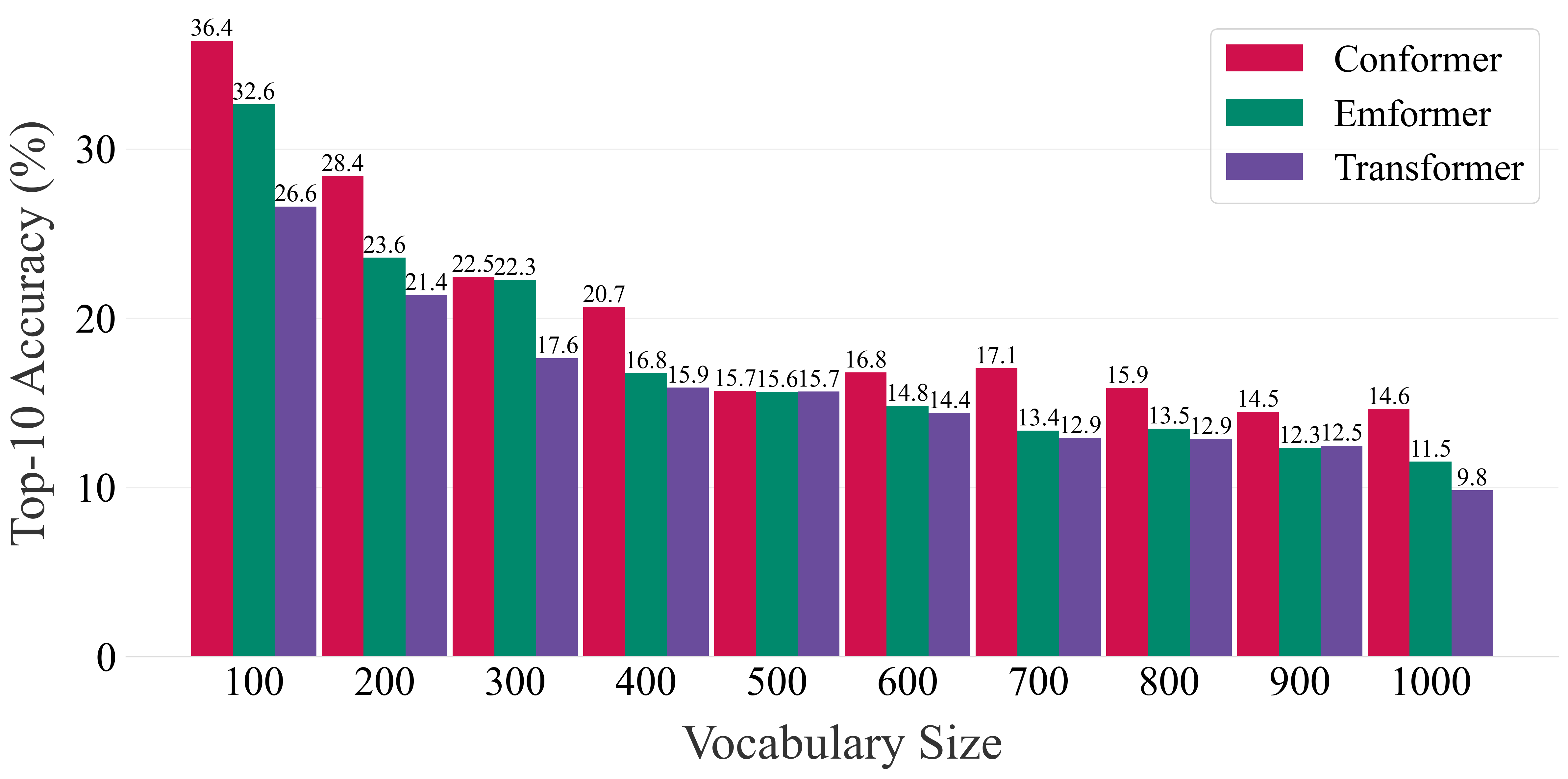}
  \caption{Ablation on different vocabulary size on different encoder architectures. \label{fig:ablation-vocab}}
\end{figure}

The ablation results on vocabulary size is depicted in Figure \ref{fig:ablation-vocab}. This result highlight the scalability of our proposed method across varying vocabulary sizes. Notably, our approach consistently outperforms competing models, demonstrating robustness even with smaller vocabulary sizes. However, as the vocabulary size increases to include more than $200$ words, we can observe a significant decline in decoding performance. This decline is primarily due to the inherent imbalance and increased scarsity in the dataset as shown in Figure \ref{fig:vocab-distribution}, where words with lower frequencies lack sufficient training data. Despite these challenges, our proposed Conformer model maintains competitive performance, achieving a top-10 accuracy of 28.4\% with a 200-word vocabulary. This result compares favorably to recent classifications of listening EEG data, which reported a top-10 accuracy of $31.4\pm1.59\%$ with a 203-word vocabulary size \cite{defossez2023decoding}. This comparison underscores the efficacy of our approach, particularly with the exclusion of grammatical words from the vocabulary, which enhances the focus on meaningful content words and improves overall performance. However, it also suggests that with a more balanced dataset, our proposed method has the potential to achieve even higher word-wise performance. This improvement would, in turn, enhance retrieval accuracy for sentence-based BCI systems on a larger scale.

\subsubsection{Visualization of word-level results} 
\begin{figure}[ht!]
  \centering
  \includegraphics[width=1\linewidth]{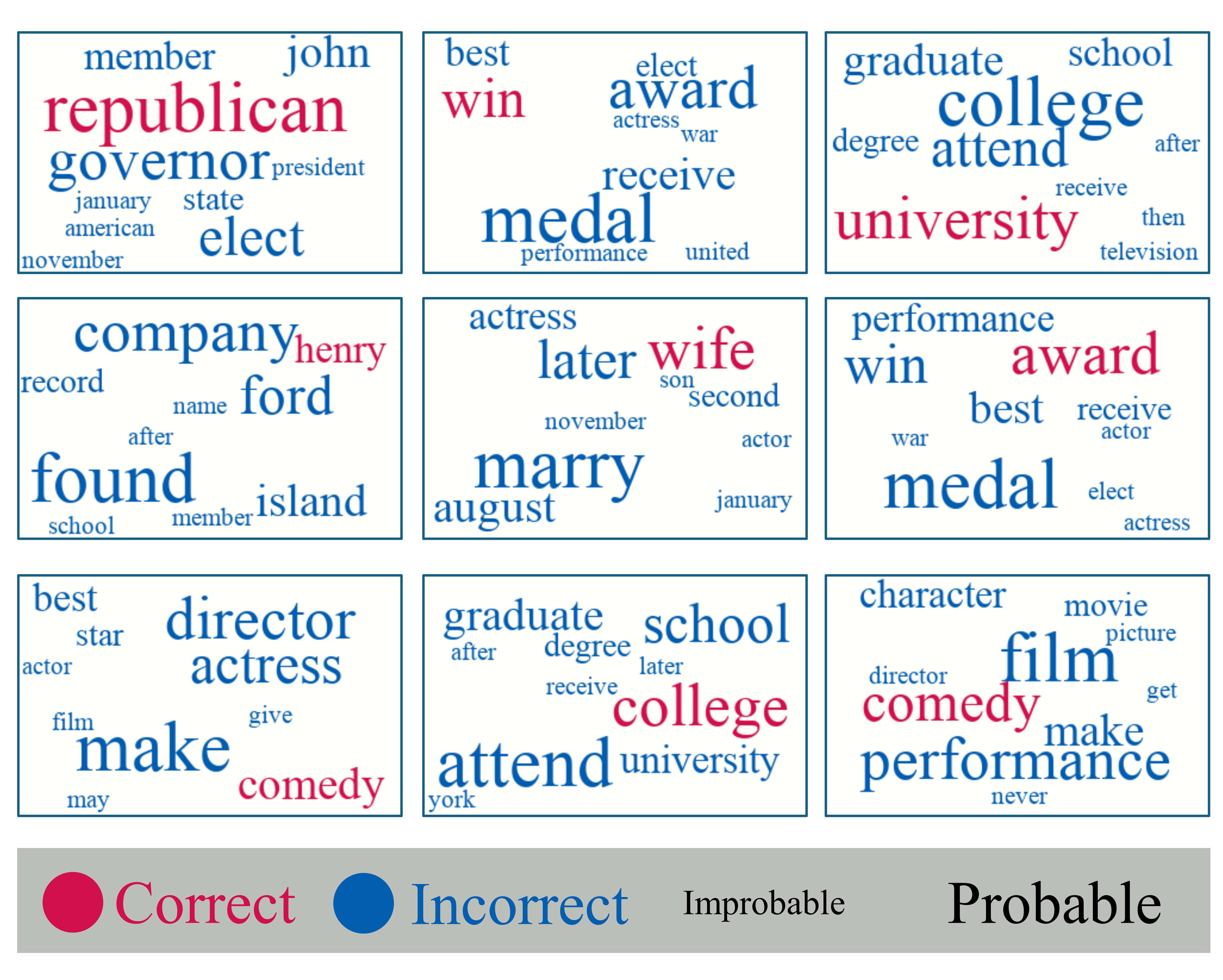}
  \caption{Visualization of top-10 prediction results from the single-word prediction using our EEG encoder. Text color indicates whether the predicted word is correct and text size is proportional to the likelihood of the model's predictions.\label{fig:wordcloud}}
\end{figure}

Figure \ref{fig:wordcloud} illustrates the top-$10$ keyword set predicted by the encoder on unseen EEG samples from the test set. After training, our model shows a strong capability of encoding EEG signals to similar concepts. For example, when predicting `university', our model also considers `graduate', `school', and `college' to be in the same keyword group, indicating the model has assigned these concepts into a close semantic representation space. We consider the clustering of meaningful concepts to support that our encoder has learned useful representation from the brain signal and has aligned these linguistic representations with language modality in the subspace. 

\subsubsection{Retrieval method and number of keyword sets} 
\begin{figure}[h!]
  \centering
  \includegraphics[width=1\linewidth]{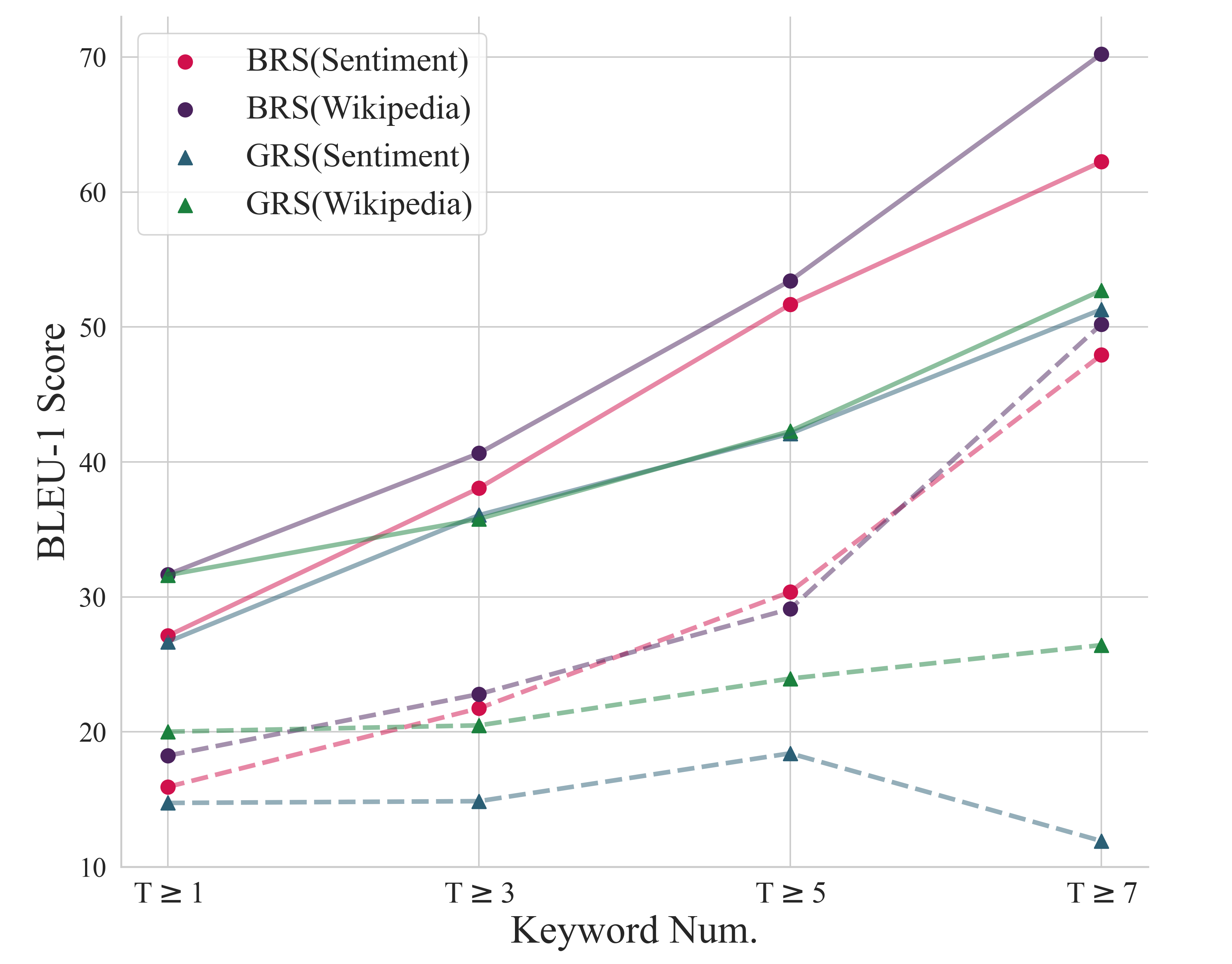}
  \caption{Comparison of retrieval relevance of BSR and GS Strategies Across Varying Numbers of EEG-Word Pairs in Sentences. Solid lines and dashed lines depict retrieval performance based on predicted keyword sets from training EEG segments and unseen EEG segments respectively.\label{Fig:decoder-ablation} }
\end{figure}

Lastly, we compare the performance of our proposed beam search retrieval method with a more straightforward greedy retrieval (GS) method and observe the retrieval performance of retrieval methods using a different number of KSs as input. The GS method is designed to use only the top-1 prediction from each keyword set for retrieving sentences from the corpus. The comparison results, presented in Figure \ref{Fig:decoder-ablation}, show that both strategies perform well on the training set as the number of keyword sets within a sentence increases. This observation suggests the potential of extracting full sentences from EEG signals using a high-accuracy EEG encoder. However, the GS method's performance significantly decreases on the test set due to its limited capacity to explore a broader array of queries, constrained by the EEG encoder's moderate top-1 prediction performance. This finding underscores the rationale for implementing a beam search-based retrieval strategy.
On the other hand, we are aware that our EEG encoder is currently constrained to the top 100 highest-frequency words, limiting the number of keywords that can be decoded from a sentence and, consequently, the application scenarios of our method. Despite this limitation, the proposed retrieval method shows improved performance when more keywords are given, even if some noise is present. These findings suggest that our method is promising and has potential for future development, particularly if the vocabulary constraint can be addressed to allow for a broader range of keywords and more accurate sentence retrieval.

\section{Conclusion}

This paper demonstrates the potential of combining an EEG encoder with a retrieval method to convert EEG signals into sentences, introducing a pioneering approach termed EEG-to-text reteival(ETER). This novel method employs a transparent EEG encoder, to learn semantic patterns from EEG data. By extracting keyword sets from unseen EEG segments, ETER enables the sentence retriever to identify the most relevant sentences from a corpus. Both quantitative and qualitative evaluations affirm the efficacy of our approach in acquiring meaningful semantic representations and retrieving relevant sentences. Our extensive experiments and ablation studies validate the approach's ability to learn patterns from unspoken EEG recordings both quantitatively and qualitatively, demonstrating that our method holds promise for providing more reliable solutions for converting EEG signals into text. Despite the achieved results,  we recognize the substantial room for future improvement. Given the exploratory nature of this research, we only employed a simple retrieval method and tested it on a limited vocabulary set. Our future work will focus on exploring more diverse datasets to continuously improve the EEG encoder design and enhance retrieval methods to accommodate larger vocabularies, thereby improving sentence retrieval accuracy on a larger scale. Additionally, collecting more EEG data at the word level will be pursued to further advance research in linguistic EEG decoding.

\bibliographystyle{ACM-Reference-Format}
\bibliography{sample-base}

\end{document}